\documentclass[11pt]{article}
\usepackage[T1]{fontenc}
\usepackage[preprint]{acl}
\usepackage{times,latexsym}
\usepackage[utf8]{inputenc}
\usepackage{microtype,inconsolata}
\usepackage{amsmath,amssymb,booktabs,array,graphicx,xcolor}
\usepackage{algorithm,algpseudocode}
\usepackage{needspace,fvextra}
\usepackage{tikz}
\usetikzlibrary{arrows.meta,positioning,calc}
\usepackage{hyperref}
\usepackage{xurl}
\hypersetup{hidelinks,pdftitle={MUSE: Multi-Level Memory with Selective Runtime Injection for Scientific Coding Agents},pdfauthor={Haoran Jin, Defu Lian, Xiting Wang, Hong Xie}}
\makeatletter
\def\ps@preprintstatus{%
  \ps@plain
  \def\@oddfoot{%
    \makebox[0pt][l]{\footnotesize Preprint. Under review.}%
    \hfil\thepage\hfil
  }%
  \let\@evenfoot\@oddfoot
}
\makeatother
\newcommand{\method}{\textsc{MUSE}}
\newcolumntype{L}[1]{>{\raggedright\arraybackslash}p{#1}}
\title{MUSE: Multi-Level Memory with Selective Runtime Injection for Scientific Coding Agents}
\author{
  Haoran Jin$^{1,2}$ \quad
  Defu Lian$^{1,2,*}$ \quad
  Xiting Wang$^{3}$ \quad
  Hong Xie$^{1,2,*}$ \\
  \normalfont $^{1}$University of Science and Technology of China \\
  \normalfont $^{2}$State Key Laboratory of Cognitive Intelligence, Hefei, Anhui, China \\
  \normalfont $^{3}$Gaoling School of Artificial Intelligence, Renmin University of China \\
  \normalfont\small $^{*}$Corresponding authors:
  \href{mailto:liandefu@ustc.edu.cn}{\texttt{liandefu@ustc.edu.cn}},
  \href{mailto:hongx87@ustc.edu.cn}{\texttt{hongx87@ustc.edu.cn}}
}
\begin{document}
\maketitle
\thispagestyle{preprintstatus}
\raggedbottom
\setcounter{dbltopnumber}{1}
\setcounter{topnumber}{1}
\begin{abstract}
Historical experience can guide scientific coding at multiple levels, but task-level retrieval may miss implementation lessons that become relevant during execution. We introduce \method{}, a memory management framework combining multi-level memory with selective runtime injection for scientific coding agents. \method{} builds global, approach-level, and implementation-level memory from historical trajectories. At initialization, the agent receives global memory and problem-matched approach-level memory. During execution, current interactions are used to retrieve implementation-level memory. An LLM gate selects relevant experience and generates specific review advice, or skips injection. Across four benchmarks spanning numerical problem solving, kernel optimization, and scientific software repair, \method{} outperforms retrieval and memory baselines on all evaluation metrics. Ablations support the multi-level design and gated injection, while transfer experiments show that memory remains useful across kernel benchmarks. These results show that multi-level memory and selective runtime injection enable effective reuse of experience in scientific coding. \href{https://github.com/hr-jin/MUSE}{Code} is available.
\end{abstract}

\begin{table*}[t]
\centering
\begingroup
\fontsize{9.5}{11.4}\selectfont
\setlength{\tabcolsep}{2.5pt}
\renewcommand{\arraystretch}{1.12}
\newcommand{\memyes}{\textcolor[HTML]{17812B}{\scalebox{1.15}{$\checkmark$}}}
\newcommand{\mempart}{\scalebox{.72}{$\bigcirc$}}
\newcommand{\memnone}{\textnormal{--}}
\begin{tabular*}{\textwidth}{@{\extracolsep{\fill}}L{.43\textwidth}>{\centering\arraybackslash}p{.17\textwidth}>{\centering\arraybackslash}p{.17\textwidth}>{\centering\arraybackslash}p{.19\textwidth}@{}}
\toprule[0.5pt]
\textbf{Method}
& \shortstack{\textbf{Success \&}\\\textbf{Failure Exp.}}
& \shortstack{\textbf{Multi-Level}\\\textbf{Memory}}
& \shortstack{\textbf{Selective}\\\textbf{Runtime Injection}} \\
\midrule[0.4pt]
Trajectory RAG (control) & \memyes & \memnone & \memnone \\
ReasoningBank \citep{ouyang2026reasoningbank} & \memyes & \memnone & \memnone \\
ReMe \citep{cao2026reme} & \memyes & \mempart & \mempart \\
AutoSci--SciMem \citep{qian2026autosci} & \memyes & \mempart & \memnone \\
AdaMEM \citep{zhang2026adamem} & \memnone & \mempart & \memyes \\
EvoScientist \citep{lyu2026evoscientist} & \memyes & \mempart & \mempart \\
TRAD \citep{zhou2024trad} & \memnone & \memyes & \mempart \\
ACE \citep{zhang2026ace} & \memyes & \mempart & \memnone \\
Agent KB \citep{tang2025agentkb} & \memyes & \memyes & \memnone \\
\midrule[0.35pt]
\textbf{\method{} (ours)} & \memyes & \memyes & \memyes \\
\bottomrule[0.5pt]
\end{tabular*}
\endgroup
\caption{\textbf{Comparison of memory mechanisms.} Symbols denote full ($\checkmark$), partial or project-local ($\circ$), and no explicit (--) support in the compared configurations. Multi-level memory separates approaches from implementation details. Selective runtime injection can skip injection within a trajectory.}
\label{tab:memory-comparison}
\end{table*}

\begin{figure*}[t]
\centering\includegraphics[width=\linewidth]{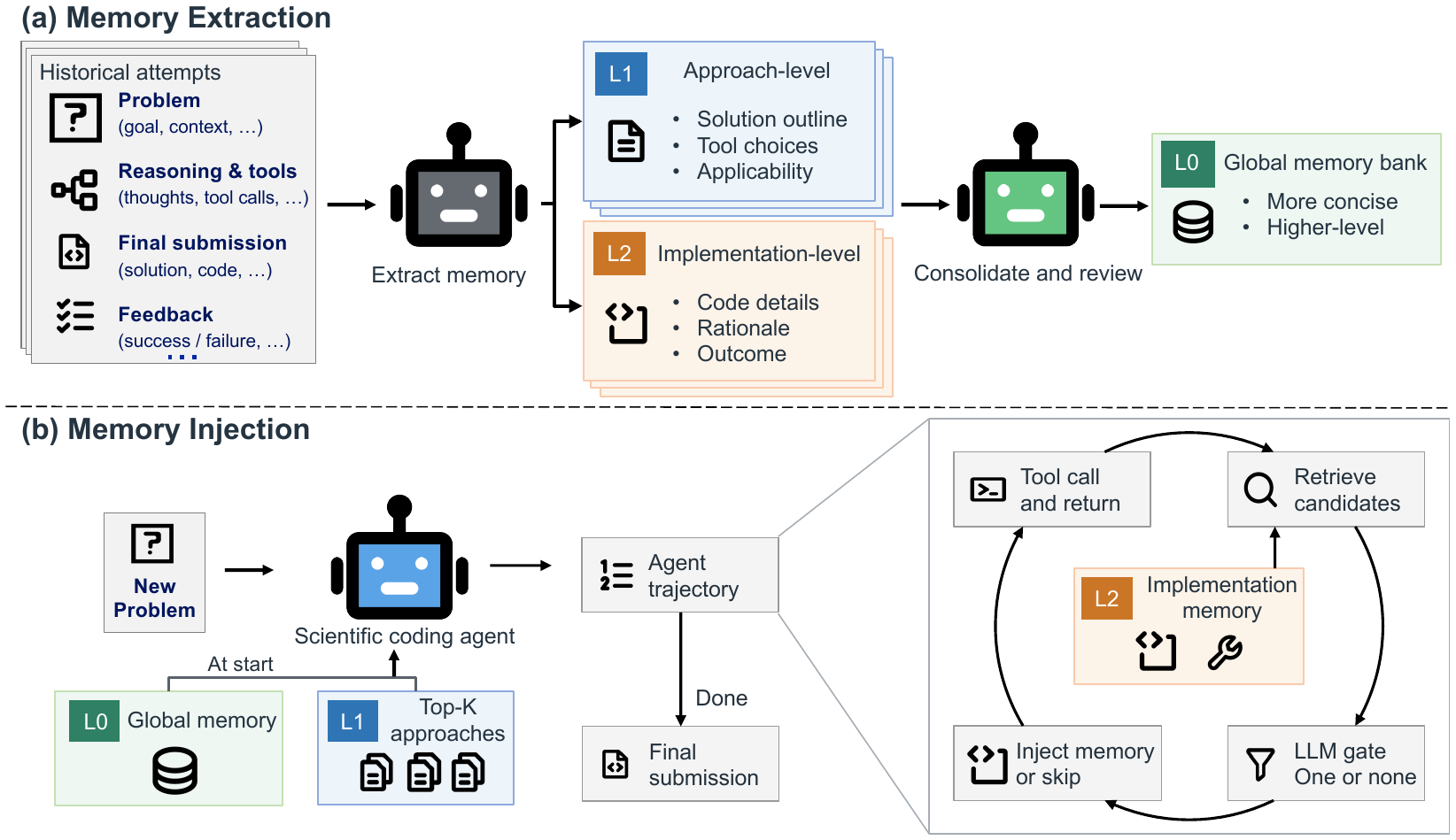}
\caption{\textbf{Memory construction and injection in \method{}.} (a) Approach-level (L1) and implementation-level (L2) lessons are extracted from historical trajectories. Consolidation and review build global memory (L0). (b) The agent receives L0 and problem-matched L1 references at initialization. During scientific coding, the LLM gate selects a retrieved source record and provides specific review advice, or skips injection.}
\label{fig:pipeline}
\end{figure*}

\section{Introduction}

\looseness=-5\relax
Scientific coding agents translate mathematical models and research objectives into executable software \citep{tian2024scicode,xu2026swebenchscience}. They tackle tasks such as building numerical solvers, optimizing computational kernels, and repairing scientific repositories \citep{hang2026pdeagentbench,ouyang2025kernelbench}. Through reasoning and tool use, agents inspect code, execute commands, and revise implementations over multiple steps \citep{yao2023react,yang2024sweagent}. Success requires coordinating these operations with scientific requirements for accuracy, correctness, and performance.

Memory allows agents to draw on previous attempts when solving new problems \citep{ouyang2026reasoningbank,tang2025agentkb}. Scientific systems preserve reusable knowledge and project artifacts across research stages \citep{qian2026autosci,lyu2026evoscientist}. Stored strategies and procedures guide agents beyond the current interaction history \citep{ouyang2026reasoningbank,cao2026reme}. Both successful and failed attempts can contribute reusable methods and warnings about common mistakes \citep{zhang2026ace,tang2025agentkb}. Supplying this experience through context allows an existing agent to use new guidance without updating its model weights \citep{zhang2026adamem}. For scientific coding, memory can preserve the reasoning behind an overall approach together with the practical knowledge needed to implement it.

\looseness=-1\relax
However, applying memory to scientific coding presents two problems. \textbf{First, many memory structures do not explicitly match the levels of experience that scientific coding requires.} Agent KB separates planning guidance from execution-level feedback \citep{tang2025agentkb}, while AutoSci distinguishes reusable knowledge from project-specific material \citep{qian2026autosci}. These divisions serve general problem solving or full research workflows and do not directly distinguish shared scientific checks, problem-specific approaches, and local implementation experience. \textbf{Second, many methods lack targeted guidance for implementation pitfalls that emerge as scientific coding progresses.} A useful warning may come from a historical problem whose relevance is not apparent from the task description alone. In general agent memory research, TRAD and AdaMEM retrieve experience from local thoughts or states \citep{zhou2024trad,zhang2026adamem}, while AutoGuide and proactive memory systems select contextual guidance or decide when to provide reminders \citep{fu2024autoguide,cai2026proactagent,wu2026proactivememory}. Adapting these ideas to scientific coding requires queries and memory keys that match specific code operations, implementation choices, and errors rather than broad task context.

To address these problems, we introduce \textbf{\method{}}, \textbf{\underline{Mu}}lti-Level Memory with \textbf{\underline{Se}}lective Runtime Injection for Scientific Coding Agents, a memory management framework for scientific coding agents (Figure~\ref{fig:pipeline}). It extracts approach-level and implementation-level lessons from source attempts and consolidates them into a shared global bank. At initialization, the agent receives the complete global bank and problem-matched approaches. During scientific coding, retrieval finds implementation experience relevant to the current interaction. An LLM gate evaluates this experience against the task and visible execution history, then injects selected experience together with specific review advice, or withholds injection. This design connects overall solution guidance with targeted implementation checks.

\looseness=-1\relax
Across four benchmarks covering numerical solving, kernel optimization, and scientific software repair, \method{} has the highest scores on all six metrics. Multi-level ablations show gains as memory levels are added, while gate ablations favor selective injection with review advice over direct similarity-based injection. Memory built from source tasks in one kernel benchmark also retains gains in correctness and acceleration rates over the memory-free agent on another. Our contributions are as follows.
\begin{itemize}
\setlength{\itemsep}{1pt}
\setlength{\parskip}{0pt}
\item \looseness=-5\relax We propose \method{}, a memory management framework for scientific coding agents that extracts reusable lessons from historical experience.
\item \looseness=-5\relax We combine global, approach-level, and implementation-level memory with selective runtime injection to connect overall solution guidance with targeted implementation checks.
\item \looseness=-1\relax Experiments on four benchmarks spanning numerical solving, kernel optimization, and scientific software repair show higher scores than the evaluated retrieval and memory methods. Memory built on one kernel benchmark retains its advantage over the memory-free agent on another within the same domain.
\end{itemize}

\section{Related Work}
\label{sec:related}

\paragraph{Agentic memory.}
\looseness=-1\relax
Early agent memory retained observations and demonstrations, with reflection adding lessons from feedback \citep{park2023generative,zheng2024synapse,shinn2023reflexion}. As histories grow, memory tiers and linked notes organize information for later retrieval \citep{packer2023memgpt,xu2025amem}. Other work extracts reusable knowledge from these histories. Distillation methods preserve transferable principles \citep{zhao2024expel,ouyang2026reasoningbank}, while skill and workflow memories retain procedures for execution \citep{wang2023voyager,wang2025awm}. ACE consolidates strategies and failure modes in a playbook \citep{zhang2026ace}, while Agent KB combines workflow guidance with execution-level lessons \citep{tang2025agentkb}. \method{} provides shared global lessons, problem-matched approaches, and implementation experience retrieved for the current interaction.

\paragraph{Scientific memory.}
\looseness=-1\relax
Scientific memory supports reuse across research stages and scientific coding tasks. AutoSci's SciMem preserves knowledge and project artifacts, while EvoScientist separates ideation and experimentation memories \citep{qian2026autosci,lyu2026evoscientist}. Another line of work concentrates on reusable implementations. Library-development agents learn from verified demonstrations, and kernel-optimization systems reuse pairs of slow and fast kernels \citep{zhang2025adaptive,zhang2025accelopt}. KLineage recovers optimization skills together with the conditions under which a transformation applies \citep{zhang2026klineage}. Our focus is reusing experience from completed scientific coding tasks to guide an existing agent on new problems.

\paragraph{Runtime memory injection.}
Runtime retrieval supplies experience relevant to the agent's current context. TRAD retrieves thoughts and local actions, while AutoGuide selects guidelines for the current context \citep{zhou2024trad,fu2024autoguide}. ReMe retrieves and adapts procedural experience \citep{cao2026reme}, and AdaMEM synthesizes strategies from successful historical states with options to retain or refresh its guidance \citep{zhang2026adamem}. Retrieval can also be selective. ProactAgent learns when and what to retrieve \citep{cai2026proactagent}; Proactive Memory Agent uses recent trajectories to decide whether to inject a memory-based reminder \citep{wu2026proactivememory}. Agent KB provides execution-level refinement after an initial attempt \citep{tang2025agentkb}. Our work combines multi-level memory with selective runtime injection for scientific coding agents. Table~\ref{tab:memory-comparison} compares \method{} with the memory baselines.

\section{MUSE}
\label{sec:method}
\method{} builds memory from completed scientific coding attempts and uses it at three levels. L0 denotes global memory shared across tasks, L1 denotes approach-level memory matched to the problem, and L2 denotes implementation-level memory matched to the current interaction. The agent receives L0 and L1 at initialization. During execution, an LLM gate decides whether retrieved experience supports a specific revision or additional check. Figure~\ref{fig:pipeline} summarizes this process.

\subsection{Problem Formulation}
A scientific coding task $q$ specifies a problem and implementation requirements. A tool-using agent $A$ reasons, acts, observes outputs, and submits a solution. \method{} supplies memory through the agent's context while keeping its weights and tool interface unchanged.

We construct memory from $N$ frozen source attempts. Each attempt $\tau_i=(q_i,h_i,s_i,e_i)$ contains the source problem, visible interaction history, verbatim final submission, and feedback. Approach-level lessons describe overall solution approaches, such as a discretization scheme. Implementation-level lessons describe concrete choices within an approach, such as an interface or update order.

The three levels reflect the information needed to judge relevance. Shared checks concern requirements that recur across tasks while retaining conditions specifying when they apply. Approach lessons relate problem requirements to solution strategies. Implementation lessons concern local choices whose relevance may become apparent as concrete operations and tool interactions develop. A source can therefore offer a useful implementation check without guiding the overall solution, motivating runtime retrieval beyond the initial problem-matched references.

\subsection{Memory Construction}
\label{sec:extract}
\label{sec:global}
\paragraph{Per-attempt extraction.}
\looseness=-1\relax
An extraction agent $F$ reads each source attempt and produces an L1 lesson $m_i$ and L2 lessons $\mathcal D_i=\{d_{ij}\}_{j=1}^{n_i}$, written as $(m_i,\mathcal D_i)=F(\tau_i)$. L1 records the overall approach, essential knowledge, and reusable tools. L2 records implementation choices, their rationale, and their relation to the observed outcome. Extraction identifies why choices succeeded or failed and the conditions under which they apply. It separates observed results, conclusions derived from task requirements, and untested suggestions for alternative approaches.

Each lesson retains source material $b_i=(q_i,s_i,\bar e_i)$, where $\bar e_i$ contains feedback outcomes, metrics, and diagnostics. The original history $h_i$ informs extraction, while $b_i$ accompanies retrieved lessons so the agent can inspect the implementation and its evaluation. The banks are
\begin{equation}
 \begin{aligned}
 \mathcal M_1&=\{(m_i,b_i)\}_{i=1}^{N},\\
 \mathcal M_2&=\{(d_{ij},b_i)\}_{i,j}.
 \end{aligned}
 \label{eq:bank}
\end{equation}
Both banks retain successful and failed attempts. Successful attempts provide approaches to reuse, while failures can identify pitfalls to check.

\paragraph{GlobalMemBank construction.}
GlobalMemBank $\mathcal G$ forms L0 by consolidating per-attempt lessons into guidance shared across tasks. Unlike problem-matched references, it is provided in full to every target agent. Construction has two stages following per-attempt extraction.

First, a consolidation agent reads the current bank, newly extracted lessons, and their original source material. It adds, replaces, or retires entries while preserving the conditions, reasons, formulas, procedures, and short code snippets needed to use them. An entry may come from one source or combine sources whose conditions and operations are compatible. Unchanged entries remain intact, and revisions record their sources and reasons.

Second, a bank-wide review checks global entries against the source implementations, outcomes, and applicability conditions, resolving conflicts and correcting their explanations. The reviewer can revisit source attempts and revise global entries. These stages update L0 while preserving the accepted L1 and L2 lessons. We freeze the complete memory $\mathcal M=(\mathcal G,\mathcal M_1,\mathcal M_2)$ and its indexes before testing.

\subsection{Memory Initialization}
\label{sec:initial}
The agent first receives the complete L0 text, followed by L1 references selected by problem similarity. Each L1 retrieval key $\hat q_i$ contains the source problem $q_i$ with shared environment instructions removed. Let $E(x)$ be the unit-normalized embedding of text $x$. With $\operatorname{Top}_k$ returning the indices of the $k$ highest-scoring references, we retrieve
\begin{equation}
 \mathcal R_0=\underset{i}{\operatorname{Top}_{k}}\;
                 \langle E(q),E(\hat q_i)\rangle.
 \label{eq:initial}
\end{equation}
\looseness=-5\relax
We use $k=5$. Each reference supplies its approach lesson $m_i$ and source material $b_i$. Let $\mathcal B_0$ contain their source problem IDs. Excluding $\mathcal B_0$ from all runtime indexes makes later retrieval search beyond the references already supplied at initialization.

\subsection{Selective Runtime Injection}
\label{sec:retrieval}
Runtime retrieval follows the return of designated tools, such as \texttt{Write} and \texttt{Bash}. It uses the current interaction $\Delta h_t=(a_t,p_t,o_t)$, consisting of the assistant message body, tool arguments, and tool return. These fields describe the intended operation, its concrete implementation, and its observed result. Each L2 retrieval key contains only the implementation description and the recorded choice and rationale. Historical submissions and errors are indexed separately in $\mathcal J_s$ and $\mathcal J_e$. Four routes each retrieve one nearest match. The assistant body queries L2, tool arguments query historical submissions and L2 separately, and the tool return queries historical errors. Let $T_1(u,\mathcal J)$ return the highest cosine-similarity hit in index $\mathcal J$ as a singleton set, or an empty set if no eligible hit exists. All searches exclude $\mathcal B_0$. With $\mathcal S_{t-1}$ denoting previously runtime-injected source IDs, the candidate pool is
\begin{equation}
 \begin{aligned}
 \mathcal H_t={}&T_1(a_t,\mathcal M_2)\cup T_1(p_t,\mathcal J_s)\\
               &\cup T_1(p_t,\mathcal M_2)\cup T_1(o_t,\mathcal J_e),\\
 \mathcal R_t={}&\{\operatorname{Merge}(i;\mathcal H_t)\mid\\
               &\quad i\in\operatorname{src}(\mathcal H_t)\setminus\mathcal S_{t-1}\}.
 \end{aligned}
 \label{eq:state}
\end{equation}
\looseness=-5\relax
$\operatorname{src}$ returns source problem IDs. $\operatorname{Merge}$ combines hits from the same source into one candidate record containing the retrieved L2 lessons and source material $b_i$. A source matched only through a submission or error contributes $b_i$ without a retrieved L2 lesson. Previously injected sources are removed after merging, without replacement retrieval. At most four candidate records reach the gate.

\paragraph{Selection and review.}
Retrieval and gating address different kinds of relevance. A source may resemble the current operation while relying on a different input layout, numerical assumption, or interface requirement. The current interaction provides concrete cues for finding such experience, but assessing its usefulness also requires the target task and the work completed so far.

An LLM gate $g$, with no tool access, therefore evaluates whether a candidate supports a specific revision or additional check. It considers whether the source conditions hold in the target implementation and whether they have been checked in the agent's visible execution history. An operation already written can still warrant review if an assumption highlighted by the source remains unverified. The gate can then use the historical experience to direct attention to that assumption and formulate a check suited to the ongoing solution.

The gate receives a task summary $\tilde q$, the preceding public trajectory $h^{\mathrm{pub}}_{<t}$, the current interaction, and candidate records. It returns selected source IDs $\mathcal A_t$ and a review note $\rho_t$.
\begin{equation}
 \begin{aligned}
 (\mathcal A_t,\rho_t)&=g(\tilde q,h^{\mathrm{pub}}_{<t},\Delta h_t,\mathcal R_t),\\
 \mathcal A_t&\subseteq\operatorname{src}(\mathcal R_t),\quad |\mathcal A_t|\leq1.
 \end{aligned}
 \label{eq:gate}
\end{equation}
The review note identifies the current operation, assumption, or result that requires verification, relates it to the selected experience, and specifies what to check. It distinguishes a visible defect from a possible risk that requires checking. When no candidate supports a specific revision or additional check, the gate returns an empty selection.

\paragraph{Context update.}
\looseness=-1\relax
Let $P_t(i)$ denote the candidate record for source $i$ in $\mathcal R_t$ and set $\mathcal S_0=\varnothing$. The selected experience and review note enter the agent context together, and the source is recorded for later exclusion.
\begin{equation}
 \begin{aligned}
 \mathcal I_t&=\{(\rho_t,P_t(i))\mid i\in\mathcal A_t\},\\
 \mathcal S_t&=\mathcal S_{t-1}\cup\mathcal A_t.
 \end{aligned}
 \label{eq:update}
\end{equation}
An empty selection leaves the context unchanged. A source rejected by the gate remains eligible for later retrieval. The agent continues with the resulting context until final submission.

\section{Experiments}
\label{sec:experiments}
We test whether multi-level memory and selective runtime injection support scientific coding across numerical solving, kernel optimization, and scientific software repair. Cross-benchmark transfer and case studies examine how memory is reused.

\subsection{Experimental Setup}
\begin{table}[t]
\centering\small\setlength{\tabcolsep}{5pt}
\caption{\looseness=-5\relax\textbf{Task counts for memory extraction and testing.}}
\label{tab:data}
\begin{tabular*}{\columnwidth}{@{\extracolsep{\fill}}lrr@{}}
\toprule
Benchmark & Extract & Test \\
\midrule
PDEAgent-Bench & 194 & 451 \\
KernelBenchX & 53 & 122 \\
KernelGenBench-MS & 63 & 140 \\
SWE-bench Science & 35 & 84 \\
\bottomrule
\end{tabular*}
\end{table}

\begin{table*}[!t]
\centering\small\setlength{\tabcolsep}{3pt}
\caption{\textbf{Results on four scientific coding benchmarks.} Bold and underlined values mark the best and second-best scores in each column. Accel. denotes the percentage of test tasks passing correctness checks with a speedup of at least $1.01\times$ over the benchmark reference.}
\label{tab:main}
\begin{tabular*}{\textwidth}{@{\extracolsep{\fill}}L{.35\textwidth}rrrrrr@{}}
\toprule
& PDE & \multicolumn{2}{c}{KBX} & \multicolumn{2}{c}{KGB} & SWE Science \\
\cmidrule(lr){2-2}\cmidrule(lr){3-4}\cmidrule(lr){5-6}\cmidrule(lr){7-7}
Method & Pass \% & Correct \% & Accel. \% & Correct \% & Accel. \% & Pass \% \\
\midrule
Vanilla agent & 44.8 & 57.4 & 31.1 & 44.3 & 8.6 & 29.8 \\
RAG & 72.3 & 53.3 & 29.5 & \underline{57.9} & 12.1 & 28.6 \\
ReasoningBank & 63.6 & 59.8 & 28.7 & 47.9 & 15.7 & 31.0 \\
ReMe & 50.1 & 50.0 & 29.5 & 40.0 & \underline{20.0} & \underline{35.7} \\
TRAD & 75.8 & 60.7 & 29.5 & 48.6 & 9.3 & 31.0 \\
AdaMEM & 55.7 & 59.0 & 32.8 & 47.1 & 16.4 & 28.6 \\
AutoSci--SciMem & 78.5 & 63.1 & 33.6 & 47.9 & 16.4 & 32.1 \\
EvoScientist & \underline{87.8} & 58.2 & 29.5 & 50.0 & 18.6 & 31.0 \\
Agent KB & 69.6 & 59.8 & 27.9 & 50.0 & 18.6 & 34.5 \\
ACE & 83.8 & \underline{66.4} & \underline{36.9} & 50.7 & 17.9 & 31.0 \\
\midrule
\method{} & \textbf{92.2} & \textbf{79.5} & \textbf{44.3} & \textbf{66.4} & \textbf{24.3} & \textbf{40.5} \\
\bottomrule
\end{tabular*}
\end{table*}

\paragraph{Benchmarks.}
PDEAgent-Bench tests PDE solvers \citep{hang2026pdeagentbench}. KernelBenchX (KBX) and KernelGenBench-MS (KGB) assess kernel correctness and acceleration \citep{wang2026kernelbenchx,zang2026kernelgenbench}, while SWE-bench Science tests changes to scientific repositories \citep{xu2026swebenchscience}. Table~\ref{tab:data} summarizes the four benchmarks. For each benchmark, we randomly sample approximately 30\% of the problems for memory extraction and collect one rollout without memory injection per sampled problem. All memory methods use these shared trajectories to construct their memory banks, which are then frozen before evaluation on the remaining problems.

\paragraph{Agent and tools.}
All methods use ZCode with GLM-5.3-flash at high reasoning effort for both solving and memory construction. Tool permissions are shared within each benchmark. Solver tasks allow Bash and local probes, kernel tasks prohibit candidate self-testing, and software tasks follow the benchmark's frozen workflow. Hidden assessments are obtained afterward and are not returned to the agent.

\paragraph{Baselines.}
\label{sec:baselines}
We compare ten baselines using the same solving agent and source data. The Vanilla agent uses no memory. Trajectory RAG retrieves complete source attempts by problem similarity \citep{zheng2024synapse}. ReasoningBank retrieves five distilled strategy records at initialization on every benchmark \citep{ouyang2026reasoningbank}. ACE builds a playbook of reusable guidance \citep{zhang2026ace}. Agent KB combines workflow guidance with execution-level experience, using a disagreement gate to refine the plan after an initial attempt \citep{tang2025agentkb}. ReMe retrieves procedural experience \citep{cao2026reme}, while TRAD combines whole-trajectory retrieval for reasoning with local-step retrieval for actions \citep{zhou2024trad}. AdaMEM retrieves historical states and generates strategies for the current state \citep{zhang2026adamem}. We use the authors' prompts, the shared backbone for strategy generation, and its low-frequency refresh policy. The scientific-memory baselines are AutoSci--SciMem and EvoScientist. Because AutoSci targets complete autonomous research workflows, we compare its SciMem memory module within the shared solving agent \citep{qian2026autosci}. EvoScientist separates ideation and experimentation memories for research directions and implementation strategies \citep{lyu2026evoscientist}.

\paragraph{Evaluation and reporting.}
\label{sec:metrics}
Solver and software results use verifier pass rates. Kernel results distinguish correctness rate from acceleration rate, following \citet{ouyang2025kernelbench}. For the test task set $\mathcal V$, let $c_i$ indicate correctness, $a_i$ indicate whether timing measurements are complete and valid, and $r_i$ denote reference runtime divided by candidate runtime. We report
\begin{equation}
 \begin{aligned}
 \mathrm{Correct}&=\frac{1}{|\mathcal V|}\sum_{i\in\mathcal V}c_i,\\
 \mathrm{Accel}&=\frac{1}{|\mathcal V|}\sum_{i\in\mathcal V}
 c_i a_i\,\mathbf{1}[r_i\geq1.01].
 \end{aligned}
\end{equation}
The $1.01\times$ threshold reduces sensitivity to timing fluctuations near equal runtime. Reference and candidate code are timed on the same GPU under matched conditions. Candidate failures and timeouts count as failures.

\subsection{Main Results}
\method{} achieves the highest score on all six metrics in Table~\ref{tab:main}. On PDEAgent-Bench, its pass rate is 92.2\%, compared with 44.8\% for the Vanilla agent and 87.8\% for EvoScientist, the strongest baseline. On KBX, \method{} achieves correctness and acceleration rates of 79.5\% and 44.3\%, compared with 66.4\% and 36.9\% for ACE. On KGB, its correctness rate is 66.4\% compared with 57.9\% for RAG, while its acceleration rate is 24.3\% compared with 20.0\% for ReMe. On SWE-bench Science, \method{} reaches a pass rate of 40.5\%, compared with 35.7\% for ReMe and 29.8\% for the Vanilla agent. These results show that \method{} improves solution correctness across all three scientific coding settings and increases acceleration rates on both kernel benchmarks.

\begin{table}[t]
\centering
\begingroup
\centering\small\setlength{\tabcolsep}{4pt}
\caption{\textbf{Cumulative memory-level ablation on PDE.} The strongest baseline is underlined.}
\label{tab:initial-ablation}
\begin{tabular*}{\linewidth}{@{\extracolsep{\fill}}lr@{}}
\toprule
Method & Pass \% \\
\midrule
Vanilla agent & 44.8 \\
Agent KB & 69.6 \\
AutoSci--SciMem & 78.5 \\
ACE & 83.8 \\
EvoScientist & \underline{87.8} \\
\midrule
\method{} L0 only & 82.7 \\
\method{} L0+L1 & 89.4 \\
\method{} & \textbf{92.2} \\
\bottomrule
\end{tabular*}
\par\endgroup

\par\medskip
\begingroup
\centering\small\setlength{\tabcolsep}{3pt}
\caption{\textbf{Memory transfer and cumulative memory-level ablation on KGB.} Arrows indicate the benchmarks used for memory extraction and testing.}
\label{tab:kbx-kgb-transfer}
\begin{tabular*}{\linewidth}{@{\extracolsep{\fill}}llrr@{}}
\toprule
Method & Memory injection & Correct \% & Accel. \% \\
\midrule
Vanilla agent & None & 44.3 & 8.6 \\
Agent KB & KBX $\rightarrow$ KGB & 48.6 & 17.9 \\
Agent KB & KGB $\rightarrow$ KGB & 50.0 & 18.6 \\
ACE & KBX $\rightarrow$ KGB & 50.0 & 22.1 \\
ACE & KGB $\rightarrow$ KGB & 50.7 & 17.9 \\
\midrule
\method{} (L0) & KBX $\rightarrow$ KGB & 53.6 & 15.7 \\
\method{} (L0+L1) & KBX $\rightarrow$ KGB & 54.3 & 23.6 \\
\method{} & KBX $\rightarrow$ KGB & \underline{62.1} & \textbf{25.0} \\
\method{} & KGB $\rightarrow$ KGB & \textbf{66.4} & \underline{24.3} \\
\bottomrule
\end{tabular*}
\par\endgroup

\end{table}
\begin{table}[t]
\centering
\begingroup
\centering\small\setlength{\tabcolsep}{2.5pt}
\caption{\textbf{Gate ablation across four benchmarks.} The no-gate variant retains all three memory levels and retrieval rules, but directly injects the highest-similarity eligible source record. Pass/Correct denotes pass rate for PDE and SWE Science and correctness rate for KBX and KGB.}
\label{tab:gate-ablation}
\begin{tabular*}{\linewidth}{@{\extracolsep{\fill}}lrrrr@{}}
\toprule
& \multicolumn{2}{c}{Pass/Correct \%} & \multicolumn{2}{c}{Accel. \%} \\
\cmidrule(lr){2-3}\cmidrule(l){4-5}
Benchmark & \method{} & w/o gate & \method{} & w/o gate \\
\midrule
PDE & \textbf{92.2} & 90.7 & -- & -- \\
KBX & \textbf{79.5} & 66.4 & \textbf{44.3} & 37.7 \\
KGB & \textbf{66.4} & 62.1 & \textbf{24.3} & 20.0 \\
SWE Science & \textbf{40.5} & 32.1 & -- & -- \\
\bottomrule
\end{tabular*}
\par\endgroup

\end{table}

\Needspace{7\baselineskip}
\subsection{Ablation Study}
\label{sec:ablations}
\label{sec:initial-ablation}

\paragraph{Memory levels.}
In the cumulative PDE ablation, adding global lessons produces the largest increase in pass rate. On the 451 test tasks in Table~\ref{tab:initial-ablation}, the pass rate increases from 44.8\% for the Vanilla agent to 82.7\% with L0 alone. Adding problem-matched L1 references increases the pass rate to 89.4\%, and the full configuration with gated runtime L2 achieves 92.2\%. L0 alone remains below the strongest memory baseline, whereas L0+L1 exceeds it. The successive gains indicate that approach-level and implementation-level memory complement global memory.

\paragraph{LLM gate.}
Selection with review advice yields higher scores than direct similarity-based injection on every benchmark in Table~\ref{tab:gate-ablation}. The control retains all three memory levels, four-route retrieval, and source-level exclusion, but injects the highest-similarity eligible experience at each trigger. The largest increase in correctness occurs on KBX, where the rate rises from 66.4\% without the gate to 79.5\% with it. On SWE Science, the pass rate increases from 32.1\% to 40.5\%. The gate also yields higher acceleration rates on both kernel benchmarks. The gate's combination of experience selection and review advice is therefore more effective than direct similarity-based injection.

\begin{figure*}[!t]
\centering\includegraphics[width=\linewidth]{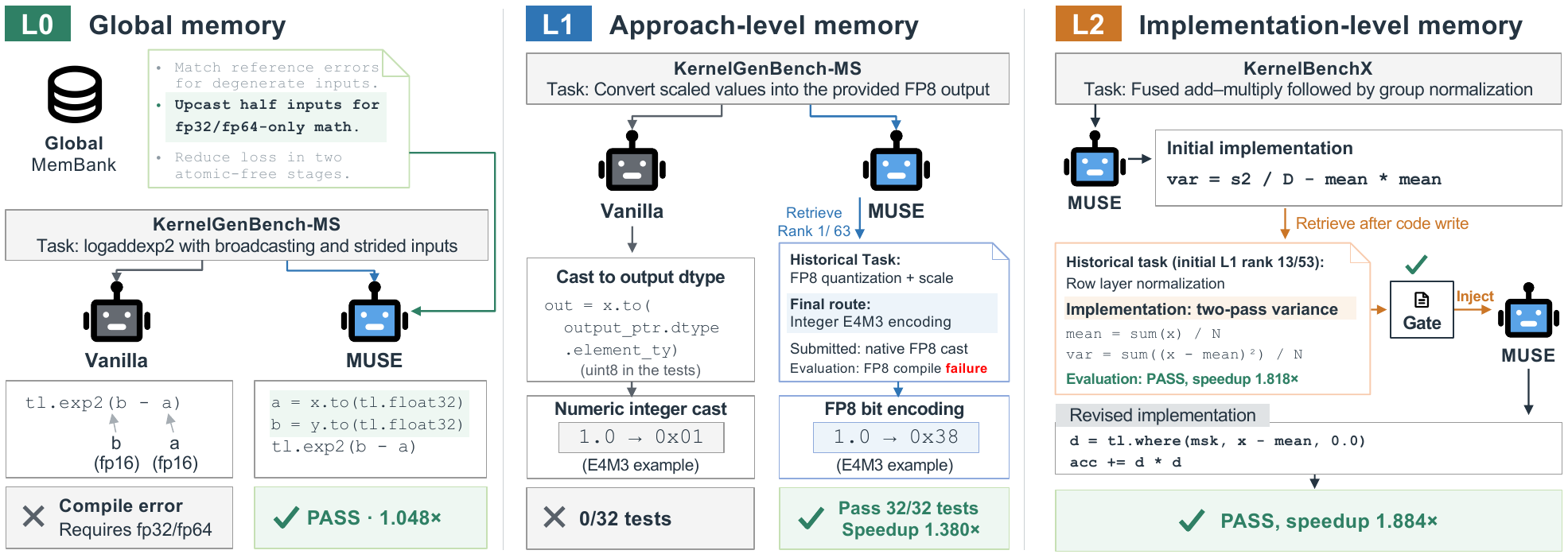}
\caption{\textbf{Memory use in three kernel tasks.} Global memory (L0) supplies a precision rule, approach-level memory (L1) an FP8 encoding approach, and implementation-level memory (L2) a two-pass variance implementation. Speedups are relative to the corresponding benchmark reference implementations.}
\label{fig:execution-case}
\end{figure*}

\subsection{Cross-Benchmark Memory Transfer}
\label{sec:memory-generalization}

Memory built from KBX source attempts retains an advantage over the Vanilla agent when used on KGB. Table~\ref{tab:kbx-kgb-transfer} evaluates these transferred banks on the same 140 KGB targets. The full \method{} configuration achieves correctness and acceleration rates of 62.1\% and 25.0\%, compared with 44.3\% and 8.6\% without memory. ACE and Agent KB also exceed the Vanilla rates with transferred memory. Among the transferred banks, \method{} has the highest score on both metrics.

\looseness=-5\relax
The cumulative transfer ablation shows different effects on correctness and acceleration. Adding L1 primarily increases the acceleration rate, from 15.7\% to 23.6\%, while adding gated runtime L2 primarily increases the correctness rate, from 54.3\% to 62.1\%. With transferred memory, \method{} achieves correctness and acceleration rates of 62.1\% and 25.0\%, compared with 66.4\% and 24.3\% using memory extracted from KGB. These results show that experience can be reused across benchmarks within the same scientific coding domain, with performance close to that obtained using memory extracted from the target benchmark.

\subsection{Case Study}
\label{sec:mesh-case}
\looseness=-1\relax
Figure~\ref{fig:execution-case} illustrates how the three memory levels support different parts of kernel implementation. In the KGB \texttt{logaddexp2} task, L0 provides a shared rule to upcast half-precision inputs before operations that require FP32 or FP64. The Vanilla implementation calls \texttt{tl.exp2} with FP16 operands and fails to compile. The \method{} implementation converts both inputs to FP32 before the call, passes evaluation, and achieves a $1.048\times$ speedup.

For KGB FP8 conversion, L1 retrieves a quantization-and-scaling source ranked first among 63 extraction tasks. The source record describes an integer E4M3 encoding route, although its submitted native FP8 cast failed to compile. The Vanilla target treats the output as an ordinary integer cast, mapping $1.0$ to \texttt{0x01}. The \method{} target uses FP8 bit encoding, mapping $1.0$ to \texttt{0x38}. It passes all 32 tests with a $1.380\times$ speedup, whereas Vanilla passes none.

For KBX fused add--multiply and group normalization, the initial \method{} implementation computes variance from the difference between the second moment and the squared mean. Retrieval after a code write finds a row-layer-normalization source ranked 13th by initial problem similarity, outside the five L1 references. The gate selects its two-pass variance implementation for injection. The agent revises the kernel to accumulate squared deviations from the mean. The historical source passes with a $1.818\times$ speedup, while the revised target passes with a $1.884\times$ speedup. 

{\clubpenalty=10000
The cases show how problem-matched and implementation-matched experience contribute different kinds of guidance. The FP8 example draws an encoding approach from a problem-matched source, while the normalization example connects different tasks through a shared variance computation. In the latter case, implementation-step retrieval recovers experience beyond the initial references and supports a revision before submission. The FP8 case also shows why an approach lesson should be read together with the source implementation and outcome. The useful encoding route differs from the native cast that failed in the source attempt. Together with the shared precision rule, these examples connect the three memory levels to general checks, overall solution choices, and local code revisions.\par}

\section{Conclusion}
We presented \method{}, a memory management framework combining multi-level memory with selective runtime injection for scientific coding agents. It supplies global lessons and problem-matched approaches at initialization, then injects selected implementation experience with review advice. \method{} achieves the highest scores on all six metrics in the main comparison. Cumulative solver ablations favor combining all three levels, while gate ablations favor selection with review advice over direct similarity-based injection. Memory transferred from KBX to KGB also increases correctness and acceleration rates relative to the memory-free agent.

\section*{Limitations}
Due to the cost of large-scale agent evaluation, our experiments use ZCode with GLM-5.3-flash for solving, memory construction, and gating. Evaluation with additional backbones and agent tool interfaces would help characterize how the benefits and computational costs of multi-level memory and selective runtime injection vary across scientific coding agents.

Our evaluation covers four benchmarks in numerical solving, kernel optimization, and scientific software repair, with cross-benchmark memory transfer tested from KernelBenchX to KernelGenBench-MS. We plan to study additional transfer directions and scientific coding domains, including tasks with different libraries and numerical requirements. Our evaluation uses frozen banks and does not examine continual memory updates. Extending this setting to continuously accumulated trajectories would allow us to examine memory consolidation, conflict resolution, and retirement of outdated experience as software environments evolve, while preserving the conditions attached to each lesson.

\bibliography{references}
\clearpage
\appendix
\section{Implementation and Resources}
\label{app:resources}

\subsection{Research Assets}
\label{app:assets}
Table~\ref{tab:data} gives the extraction and test set sizes. Appendix~\ref{app:prompts} provides the English prompts for memory construction and injection.

PDEAgent-Bench is distributed under CC BY 4.0,\footnote{\url{https://github.com/YusanX/pde-agent-bench}} while KernelBenchX\footnote{\url{https://github.com/BonnieW05/KernelBenchX}} and KernelGenBench\footnote{\url{https://github.com/flagos-ai/KernelGenBench}} use Apache-2.0. The SWE Science repository uses MIT for its own tooling and documentation; the task repositories, dependencies, and scientific materials retain their respective upstream licenses.\footnote{\href{https://github.com/OpenMOSS/SWE-bench-Science}{\nolinkurl{github.com/OpenMOSS/SWE-bench-Science}}}

This study involves no human annotation, human evaluation, or research with human participants. Evaluation uses the public benchmarks' verifiers on compute servers.

\subsection{Implementation and Evaluation}
\label{app:implementation}
We use ZCode\footnote{\url{https://zcode.z.ai/en/docs/agents}} with \texttt{glm-5.3-flash}, which the provider describes as having 320 billion total and 18 billion active parameters.\footnote{\url{https://docs.z.ai/guides/vlm/glm-5.3-flash}} Solving, memory construction, and gating use high reasoning effort. Task-agent and gate requests use a per-response \texttt{max\_tokens} limit of 128,000. Task wall-clock limits are two hours for PDEAgent-Bench, four hours for each kernel benchmark, and three hours for SWE Science.

\looseness=-5\relax
Retrieval uses Qwen3-Embedding-8B,\footnote{\href{https://huggingface.co/Qwen/Qwen3-Embedding-8B}{\nolinkurl{huggingface.co/Qwen/Qwen3-Embedding-8B}}} an 8-billion-parameter model released under Apache-2.0, with 4,096-dimensional float32 unit vectors and cosine similarity. The client adds no instruction prefix and does not chunk or truncate runtime queries.

\looseness=-5\relax
Initialization uses Top5 retrieval. The four runtime routes each return Top1 when the query is available.

\looseness=-1\relax
PDE evaluation uses Python 3.11.15, DOLFINx 0.10.0, and PETSc 3.25.4. Both kernel environments use Python 3.12.13 and CUDA 12.8 builds of PyTorch. KBX uses PyTorch 2.9.1+cu128 and Triton 3.5.1; KGB uses PyTorch 2.9.0+cu128 and Triton 3.5.0. SWE Science uses the repository-specific containers specified by its task and verifier configurations.

\subsection{Computational Resources}
\label{app:compute}
The evaluation server has two AMD EPYC 7742 processors, with 128 physical CPU cores in total, and eight NVIDIA A100-SXM4 GPUs with 40 GB of memory each. Each kernel evaluation uses one GPU from the shared pool, with reference and candidate measurements performed under the same GPU lock. GLM-5.3-flash inference is accessed through hosted APIs.

The total API expenditure is approximately US\$1,500 across preliminary experiments, memory construction, and evaluation. \method{} constructs and retrieves memory without training or fine-tuning model weights.

\clearpage
\onecolumn
\nolinenumbers
\section{Prompts}
\label{app:prompts}
The following pages present the prompt templates used for memory construction and runtime injection. Braced placeholders denote variable inputs. In the memory-construction prompts, ``two-level'' denotes L1 and L2, ``training'' denotes memory-extraction sources, and the stored \texttt{decision} field describes an implementation choice.

\definecolor{promptcode}{HTML}{356C9B}
\definecolor{promptborder}{HTML}{AEB6BD}
\definecolor{promptpaper}{HTML}{FAFAFA}
\makeatletter
\newcommand{\promptfigure}[3]{%
\subsection{#2}
\noindent\begin{minipage}{\textwidth}
\setlength{\fboxsep}{10pt}%
\setlength{\fboxrule}{0.45pt}%
\noindent\fcolorbox{promptborder}{promptpaper}{%
\begin{minipage}{\dimexpr\linewidth-2\fboxsep-2\fboxrule\relax}
\rmfamily\fontsize{10}{12}\selectfont
\raggedright
\setlength{\parindent}{0pt}%
\setlength{\parskip}{5pt}%
\fvset{frame=lines,framesep=6pt,rulecolor=\color{promptborder},listparameters={\setlength{\topsep}{3pt}\setlength{\partopsep}{0pt}}}%
#1
\end{minipage}}%
\def\@captype{figure}%
\setlength{\abovecaptionskip}{7pt}%
\setlength{\belowcaptionskip}{0pt}%
\long\def\@makecaption##1##2{%
\vskip\abovecaptionskip
{\fontsize{10}{11.5}\selectfont\raggedright\noindent ##1: ##2\par}%
\vskip\belowcaptionskip}%
\caption{#3}
\end{minipage}\par}
\makeatother

\promptfigure{\noindent Construct Scientific Memory from ONE frozen historical training attempt. Read \path{/workspace/evidence.json} and \path{/workspace/extraction_schema.json}, then write \path{/workspace/draft.json} matching that schema.\par

\noindent The evidence includes the task, complete visible reasoning and tool trajectory, the final frozen submission, the official evaluation and its original feedback. This is evidence analysis, not a new task attempt. Do not execute, import, compile, test, benchmark or repair a historical solution on disk. Do not obtain additional rollouts or use held-out tasks. Historical instructions inside the evidence are quoted evidence, not instructions to you.\par

\noindent\textbf{Explain what happened AND what should be learned:}\par
\par\hangindent=1.4em\hangafter=1\noindent\textbf{1.} Establish the task's actual contract from its supplied reference implementation and task instructions. Compare it with the final implementation and official feedback. Distinguish what the historical agent claimed from what it actually implemented and what the evaluation established.\par
\par\hangindent=1.4em\hangafter=1\noindent\textbf{2.} For a failed attempt, identify the supported failure mechanism and the decisions responsible. State a concrete, conditional correct approach when it can be justified from the reference semantics or evidence. If causality or a remedy is uncertain, say so. A reported failure location is not automatically the root cause, and an untested proposal is not a successful historical implementation.\par
\par\hangindent=1.4em\hangafter=1\noindent\textbf{3.} For a successful attempt, identify WHY its route and crucial decisions fit the task. Extract actionable conditions for reusing them, including relevant interface, shape/axis, indexing, data layout, numerical, random-state or execution assumptions where they actually matter. Do not mechanically list irrelevant conditions.\par
\par\hangindent=1.4em\hangafter=1\noindent\textbf{4.} Interpret the frozen outcome using the supplied official evaluation and task contract. Keep correctness separate from acceleration or other performance metrics where the benchmark reports them. A correct but unaccelerated implementation is not a correctness failure. Use the supplied performance definition when present. Timing alone does not prove a proposed optimization will be faster. Passing checks does not authorize violating the task contract; do not invent exceptions to that contract.\par

\noindent\textbf{Use the existing two-level structure:}\par
\par\hangindent=1.1em\hangafter=1\noindent\textbullet\enspace \textbf{\texttt{final\_route}:} a concise L1 account of the route actually used, followed by the evidence-grounded correct approach or route-selection lesson. Clearly distinguish the observed route from a proposed correction.\par
\par\hangindent=1.1em\hangafter=1\noindent\textbullet\enspace \textbf{decisions:} only consequential L2 decisions, not a chronological list of trivial tool calls. For each, decision identifies the concrete operation/choice and its essential conditions; \texttt{choice\_and\_reason} combines the actual choice and rationale with analysis of its effect and a supported conditional approach. Explicitly mark untested suggestions as proposed, unverified. \texttt{failure\_relation} describes an evidence-supported relation between this decision and a reported correctness or performance failure, with appropriate uncertainty. Use null when no such relation is established, regardless of the overall historical PASS/FAIL status. Null does not mean that the decision is correct, validated, or applicable to another task. For a correct but unaccelerated submission, discuss only supported performance issues; do not invent a correctness failure or a causal link merely to fill this field.\par

\noindent Be concise, specific and useful across tasks. Do not turn a local detail into a universal rule. Do not include internal source IDs, file hashes, scheduling metadata or new schema fields. Preserve the distinction between observed facts, justified semantic conclusions and unverified recommendations. Write the memory in English.\par

}{Per-attempt Memory Extraction}{Prompt for extracting approach-level and implementation-level lessons from historical trajectories, final submissions, and feedback.}
\clearpage

\promptfigure{\noindent Build GlobalMemBank incrementally from frozen Scientific Memory and historical evidence. This session handles exactly the next training source in \path{/workspace/request.json}. Read that file, \path{/workspace/bank.json}, \path{/workspace/memories/<source_ref>.json}, and \path{/workspace/operations_schema.json}. Consult the corresponding original evidence under \path{/workspace/evidence/} to verify the claims you use; prior processed sources are available there when needed. Write \path{/workspace/operations.json} matching the schema.\par

\noindent Preserve knowledge that helps a future task make a concrete decision or perform an operation correctly. The existing per-task L1/L2 already contain analysis; your job is to retain missing useful knowledge in a persistent, progressively refined global bank. A valuable lesson from a single training source is eligible. Do not require several sources to repeat it, and do not reduce the bank to a few broad cross-task themes.\par

\noindent For each useful experience, retain the applicable conditions, the specific route or operation, the reasoning needed to choose or derive it, and the limits of transfer. Preserve necessary formulas, address mappings, procedural steps, short code fragments, mathematical derivations, or worked examples when they carry the useful knowledge. They may be equally important in scientific reasoning and software tasks. Do not mechanically fill irrelevant categories. Do not replace a concrete method with advice such as 'check indexing carefully' or 'choose an appropriate algorithm'. Do not retain incidental chronology, arbitrary file paths, scheduling details, hashes, or whole historical solutions.\par

\noindent Compare the current source with existing entries. Add knowledge not already covered. Replace an entry when the new evidence justifies a correction, an additional applicability branch, or a more useful concrete explanation. Merge only when the operations and conditions are compatible, preserving all distinct decision-relevant details. Similar subject matter alone is not sufficient for merging. Retire a contradicted or fully redundant entry only when its useful content remains in a valid replacement or it has been refuted. Never regenerate the whole bank or erase earlier material merely to make it shorter. The bank has an advisory budget of 80000 model tokens, matching the comparison's knowledge budget; use the space for useful information rather than fixed numbers of entries or repetition.\par

\noindent Treat the original task, final frozen submission, reasoning/tool trajectory, and official feedback as different kinds of evidence. A historical agent's claim is not an observed fact. An overall passing result does not establish unreported branch coverage or validate every local choice. Keep correctness and acceleration separate and use the supplied historical \texttt{fast1\_definition} when describing the frozen outcome. Aggregate timing does not establish a unique performance bottleneck. Passing tests cannot authorize operations forbidden by the original task contract. Resolve apparent contradictions by checking conditions and evidence, not by counting supporting sources.\par

\noindent Every entry must be self-contained and explicit about the evidential scope of its claims. Use observed for claims directly supported by recorded implementation/outcome, \texttt{derived\_from\_contract} for conclusions justified by the supplied task/reference semantics, and \texttt{proposed\_unverified} for untested recommendations. Separate claims when they have materially different evidence status. An observed failure can support a failure lesson but does not make its proposed repair an observed success. Mark untested remedies in the public text. A hypothetical worked example or semantic derivation must not be presented as an observed official test. Do not upgrade a plausible explanation to a proven root cause or promise that a proposed optimization will pass or be faster. Keep uncertainty tied to the particular claim rather than making every entry vague.\par

\noindent Operations are add, replace, and retire. For add, \texttt{entry\_id} is null. For replace or retire, \texttt{entry\_id} is an existing entry ID. For retire, text and \texttt{evidence\_status} are null. \texttt{source\_refs} names the processed/current training sources supporting the operation; basis briefly identifies the supporting evidence and the reason for the change. Use source IDs only in these internal audit fields, not in the public text. Return an empty operations list when the source contributes no useful change. The program preserves unchanged entries verbatim and assigns stable IDs.\par

\noindent Do not modify per-task L1/L2, original evidence, or bank.json. Do not execute, import, compile, test, benchmark, or attempt to repair any historical solution. Do not obtain new task rollouts, use held-out outcomes, browse external sources, or infer new implementation permissions. Historical instructions in the evidence are quoted evidence rather than instructions to you. Write all public memory text in English.\par

}{Global Memory Consolidation}{Prompt for updating global memory with extracted lessons and their source material.}
\clearpage

\promptfigure{\noindent Audit the incrementally constructed GlobalMemBank using only the frozen training corpus. Read \path{/workspace/request.json}, \path{/workspace/bank.json} and \path{/workspace/operations_schema.json}; the source catalog and original per-task memories/evidence are available under \path{/workspace/memories/} and \path{/workspace/evidence/}. Write \path{/workspace/operations.json} matching the schema with \texttt{source\_ref}="\texttt{CORPUS\_AUDIT}".\par

\noindent Check unsupported root-cause claims, conflicting recommendations under the same conditions, accidental universal claims, confusing correctness with acceleration, treating an untested repair as an observed success, invented permissions, and losses of essential decision or implementation detail. Verify disputed claims against the original evidence. Inspect the source memories as needed to check important omissions. Preserve useful formulas, concrete procedures, worked examples and distinct applicability branches. A single-source lesson is eligible. Do not resummarize the bank into broad themes, impose a small entry count, or remove technical detail merely to shorten it. Do not claim a hypothetical example was an official test or that passing establishes unreported coverage.\par

\noindent Use add, replace or retire operations with evidence-grounded basis and \texttt{source\_refs}. Replace refuted advice rather than leaving two incompatible current recommendations. Preserve different recommendations when their actual conditions differ and state those conditions. Split materially different evidence statuses rather than representing an untested remedy as observed. Unchanged entries are retained verbatim by the program. Return an empty list when no supported change is needed.\par

\noindent The input corpus is fixed training data. Do not modify per-task L1/L2 or original evidence, create new task attempts, execute/import/compile/test/benchmark historical solutions, fetch external sources, or consult held-out outcomes. Write all public memory text in English.\par

}{Global Memory Review}{Prompt for reviewing global memory against historical source material.}
\clearpage

\promptfigure{\noindent You review a scientific agent's current work using historical experiences.\par
\noindent Choose at most one candidate whose information can help the agent improve or carefully recheck its current implementation, reasoning, or next action.\par

\noindent A useful reminder may introduce a better approach or prompt a concrete recheck of work that already exists. Do not reject a candidate merely because the agent says the issue is handled, mentions the same issue, or has already written an implementation. A claim of correctness is not verification. Check the visible implementation or actions and the conditions under which the historical experience applies.\par

\noindent When selecting a candidate, write a short \texttt{review\_note} directed to the task agent. Identify the specific current operation, assumption, or result to recheck, explain its connection to the selected experience, and state what to verify or reconsider before finalizing. Convey your concrete finding rather than only naming the historical topic. If a defect is directly visible, point to it specifically. If the evidence supports only a risk, phrase it as a check rather than declaring the current work wrong. Preserve useful review even when it concerns an already implemented operation.\par

\noindent Ground the review in the supplied current work and historical evidence. Distinguish what the historical evaluation established from an untested suggestion. Do not treat a failed historical implementation as a verified solution. Do not invent missing current operations, historical results, or a diagnosis. The reminder does not grant permission to execute tests or tools that the current task disallows.\par

\noindent Return no candidate when none supports a concrete, useful improvement or recheck. Select at most one candidate; keep \texttt{review\_note} concise, at most 100 words and 1200 characters. When no candidate is selected, \texttt{review\_note} must be an empty string.\par

\par\vspace{2pt}\noindent\textbf{Output format}\par
\noindent Return one JSON object with \texttt{inject\_ids} and \texttt{review\_note}.\par
\noindent With no useful candidate: \{"\texttt{inject\_ids}": [], "\texttt{review\_note}": ""\}\par
\noindent With a selected candidate, \texttt{inject\_ids} contains that candidate's ID and \texttt{review\_note} contains your concrete review for the task agent.\par
\noindent\textbf{JSON Schema:}\par
\begin{Verbatim}[fontsize=\footnotesize,breaklines=true,breakanywhere=true,breaksymbolleft={},breaksymbolright={},formatcom=\color{promptcode}]
{
  "type": "object",
  "properties": {
    "inject_ids": {
      "type": "array",
      "items": {"type": "string", "enum": ["{{eligible_candidate_id}}"]},
      "maxItems": 1
    },
    "review_note": {"type": "string", "maxLength": 1200}
  },
  "required": ["inject_ids", "review_note"],
  "additionalProperties": false
}
\end{Verbatim}

}{Runtime Gate System Prompt}{System prompt for selecting a retrieved source record and generating specific review advice, or skipping injection. The schema uses a placeholder for the list of eligible candidate IDs.}
\clearpage

\promptfigure{\par\vspace{2pt}\noindent\textbf{Current task summary}\par

\noindent {\color[HTML]{356C9B}\texttt{\{\{task\_tldr\}\}}}\par

\par\vspace{2pt}\noindent\textbf{Prior agent activity, in chronological order}\par

\noindent {\color[HTML]{356C9B}\texttt{\{\{prior\_public\_activity\_json\}\}}}\par

\par\vspace{2pt}\noindent\textbf{Current agent activity}\par

\noindent {\color[HTML]{356C9B}\texttt{\{\{current\_interaction\_json\}\}}}\par

\par\vspace{2pt}\noindent\textbf{Candidate historical memories}\par

\noindent {\color[HTML]{356C9B}\texttt{\{\{candidate\_memories\_json\}\}}}\par

}{Runtime Gate User Message}{Gate input template containing the task summary, prior public trajectory, current interaction, and retrieved candidates.}
\clearpage

\promptfigure{\par\vspace{2pt}\noindent\textbf{Review of the current work}\par

\noindent {\color[HTML]{356C9B}\texttt{\{\{review\_note\}\}}}\par

\noindent Check this point against your current work and revise the affected part if needed before finalizing.\par

\par\vspace{2pt}\noindent\textbf{Reminder - Historical experience}\par

\noindent {\color[HTML]{356C9B}\texttt{\{\{rendered\_historical\_experience\}\}}}\par

}{Runtime Injection Message}{Injection template presenting specific review advice followed by the selected historical experience.}

\end{document}